\documentclass[10pt,twocolumn,letterpaper]{article}

\usepackage{cvpr}              % To produce the CAMERA-READY version
\usepackage{amssymb}
\usepackage{amsmath}
\usepackage{multirow}

\definecolor{cvprblue}{rgb}{0.21,0.49,0.74}
\usepackage[pagebackref,breaklinks,colorlinks,allcolors=cvprblue]{hyperref}

\def\paperID{17465} % *** Enter the Paper ID here
\def\confName{CVPR}
\def\confYear{2025}

\title{Overcoming Semantic Dilution in Transformer-Based Next Frame Prediction}

\author{
Hy Nguyen, Srikanth Thudumu, Hung Du, Rajesh Vasa, Kon Mouzakis\\
Applied Artificial Intelligence Institute, Deakin University\\
Burwood, Victoria, Australia\\
}

\begin{document}
\maketitle
\begin{abstract}
Next-frame prediction in videos is crucial for applications such as autonomous driving, object tracking, and motion prediction. The primary challenge in next-frame prediction lies in effectively capturing and processing both spatial and temporal information from previous video sequences. The transformer architecture, known for its prowess in handling sequence data, has made remarkable progress in this domain. However, transformer-based next-frame prediction models face notable issues: (a) The multi-head self-attention (MHSA) mechanism requires the input embedding to be split into $N$ chunks, where $N$ is the number of heads. Each segment captures only a fraction of the original embedding’s information, which distorts the representation of the embedding in the latent space, resulting in a semantic dilution problem; (b) These models predict the embeddings of the next frames rather than the frames themselves, but the loss function based on the errors of the reconstructed frames, not the predicted embeddings -- this creates a discrepancy between the training objective and the model output. We propose a Semantic Concentration Multi-Head Self-Attention (SCMHSA) architecture, which effectively mitigates semantic dilution in transformer-based next-frame prediction. Additionally, we introduce a loss function that optimizes SCMHSA in the latent space, aligning the training objective more closely with the model output. Our method demonstrates superior performance compared to the original transformer-based predictors.
\end{abstract}    
\section{Introduction}
\label{sec:intro}

Humans can anticipate short-term future events using visual information, a capability crucial for tasks such as autonomous driving \cite{wu2020motionnet}, action recognition \cite{kong2022human}, motion prediction \cite{kong2022human}, and anomaly detection \cite{lu2019future}. To enable machines to develop similar abilities, the task of Video Frame Prediction (VFP) has been extensively studied, involving the prediction of future video frames from recent past frames. 

VFP is challenging due to the complexity of real-world video dynamics and the inherent uncertainty of future events \cite{zhu2024video}. Capturing both spatial and temporal information effectively is essential \cite{lin2020self}. Before Transformers \cite{vaswani2017attention}, approaches typically combined sequence models (e.g., LSTM, RNN, GRU) with CNNs to extract spatio-temporal features from past frames \cite{lin2020self, wang2017predrnn, wang2018eidetic, wu2021motionrnn}. However, these models struggle with long-term dependencies, suffer from vanishing gradients, and are computationally expensive and error-prone over longer sequences \cite{bengio1993problem}, \cite{lin2020self}, \cite{ye2022vptr}. In contrast, Transformer-based methods using Multi-head Self-Attention (MHSA) are more effective at handling long-range dependencies and allow for efficient parallel processing \cite{vaswani2017attention, lin2020self, arnab2021vivit, ye2022vptr, ning2023mimo, zhu2024video}.

\begin{figure}[!ht]
    \centering
    \includegraphics[width=1\linewidth]{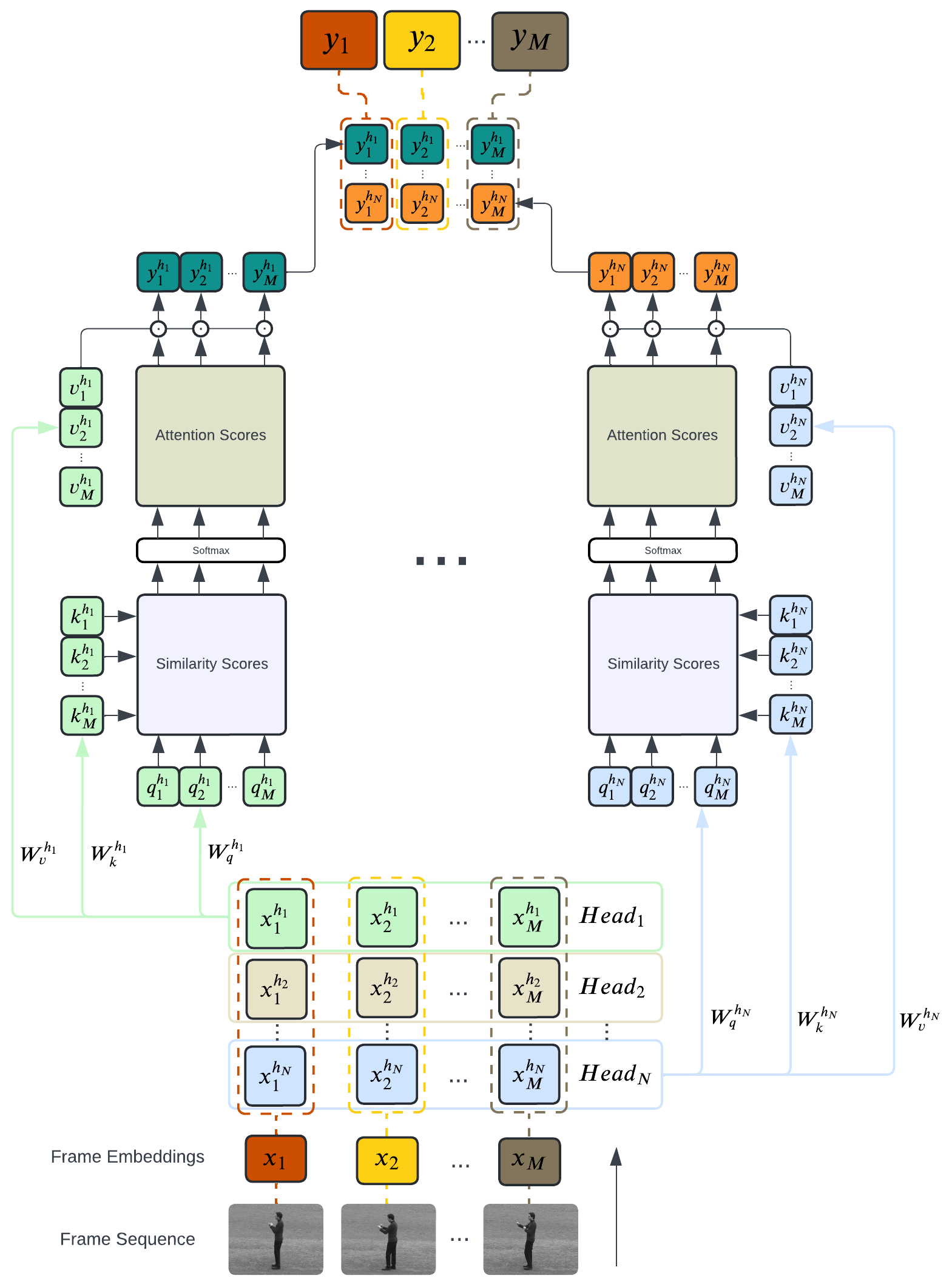}
    \caption{Multi-head Self-attention (MHSA) block in VFP divides the input embedding $x_i$ into $N$ chunks $\{x^{h_1}_i, x^{h_2}_i, ..., x^{h_N}_i\}$ where $N$ is the number of attention heads. This can distort the learned embeddings, leading to semantic dilution of the frame representation.}
    \label{fig:semantic_dilution}
\end{figure}

However, MHSA requires the input embedding to be divided into multiple chunks (corresponding to the number of attention heads), which can distort the learned latent space and dilute semantic information, reducing prediction accuracy. Figure \ref{fig:semantic_dilution} illustrates how the input embedding is partitioned in the MHSA implementation. Although the Transformer-based VFP systems introduce additional modules to enhance prediction performance, such as the Local Spatio-Temporal block \cite{ning2023mimo}, Self-attention Memory (SAM) \cite{lin2020self}, or Temporal MHSA \cite{ye2022vptr}, etc, the core MHSA mechanism remains unchanged, allowing the issue of semantic dilution to persist. Furthermore, these VFP systems do not directly predict the next frame; instead, they predict the embedding of that frame. This requires a decoding step to reconstruct the predicted frame from its embedding. However, the loss function used to train these VFP systems is based on the error of the reconstructed frame, not the predicted embedding, leading to a discrepancy that can hinder effective model learning. For example, a common loss function used in VFP systems is structured as follows \cite{lin2020self, ning2023mimo, girdhar2021anticipative}:
\begin{equation}
    \mathcal{L} = \mathcal{L}_1(X,\hat{X}) + \mathcal{L}_2(X, \hat{X}) + \mathcal{P}/\mathcal{S}(X, \hat{X})
    \label{eq:common_VFP_loss}
\end{equation}
where $X$ is the ground truth frame, $\hat{X}$ is the predicted frame, $\mathcal{L}_1$ represents the L1 loss (Mean Absolute Error, MAE), $\mathcal{L}_2$ represents the L2 loss (Mean Squared Error, MSE), and $\mathcal{P}/\mathcal{S}$ denotes Perceptual loss or Structural loss (such as image gradient loss \cite{mathieu2015deep}, Charbonnier loss \cite{charbonnier1994two}, Census loss \cite{meister2018unflow}, LPIPS loss \cite{zhang2018unreasonable}, etc). As seen, this loss function depends on the predicted frame $\hat{X}$ rather than the predicted embedding. While this approach can still be effective, it does not fully align with the output of VFP systems, which predict embeddings. This misalignment introduces challenges for model optimization such as slower convergence, suboptimal learning, and gradient mismatch. How to address the issue of semantic dilution and eliminate the misalignment in the learning objective of Transformer-based VFP systems?

To tackle these challenges, we introduce a semantic-preserving module called Semantic Concentration Multi-Head Self-Attention (SCMHSA) for Transformer-based VFP systems. This module allows the query, key, and value matrices for each attention head to be computed using the entire input embedding, thereby mitigating the semantic dilution issue in standard MHSA blocks. However, this approach increases the dimensionality of each head, making it more challenging to train the heads to support each other integratively. To address this, we propose a loss function that ensures each head, while using the complete input embedding, focuses on distinct semantic aspects, thereby facilitating more efficient convergence. Unlike existing methods that operate in the pixel space (based on the reconstructed frame), our loss function is designed to work in the embedding space (based on the predicted embedding), aligning the training objective more closely with the VFP output. This alignment also enhances the efficiency of the autoregressive mechanism within the SCMHSA block, as the autoregression operates in the embedding space. By predicting the embedding of the next frame instead of performing a full pixel-level reconstruction, our approach effectively captures essential features, making it particularly well-suited for tasks such as anomaly detection in object tracking, where the focus is on identifying significant deviations or irregularities rather than requiring complete visibility of the next frame. We evaluate our method on four different datasets: KTH \cite{schuldt2004recognizing}, UCSD \cite{5539872}, UCF Sports \cite{soomro2015action}, and Penn Action \cite{zhang2013actemes}. 

Our contributions can be summarized as follows:
\begin{enumerate}
    \item We introduce a Semantic Concentration Multi-Head Self-Attention (SCMHSA) block that preserves the semantics in the embeddings of the input frame sequence for VFP systems.
    \item We present a new loss function based on the embedding space rather than the pixel space, effectively aligning with the VFP output and ensuring that each head can focus on distinct semantic aspects.
    \item Through empirical evaluations, we demonstrate that our proposed method outperforms existing Transformer-based VFP techniques in terms of prediction accuracy.
\end{enumerate}

% \begin{figure*}
%   \centering
%   \begin{subfigure}{0.68\linewidth}
%     \fbox{\rule{0pt}{2in} \rule{.9\linewidth}{0pt}}
%     \caption{An example of a subfigure.}
%     \label{fig:short-a}
%   \end{subfigure}
%   \hfill
%   \begin{subfigure}{0.28\linewidth}
%     \fbox{\rule{0pt}{2in} \rule{.9\linewidth}{0pt}}
%     \caption{Another example of a subfigure.}
%     \label{fig:short-b}
%   \end{subfigure}
%   \caption{Example of a short caption, which should be centered.}
%   \label{fig:short}
% \end{figure*}

\section{Related Work}
\label{sec:related_work}
Before the introduction of the Transformer model \cite{vaswani2017attention}, VFP systems primarily relied on sequence models such as LSTM, RNN, and GRU for temporal information, and image-to-vec models such as CNNs and Autoencoders for spatial information. These temporal and spatial models were often combined to process historical frame sequences and predict the next frame. ConvLSTM \cite{shi2015convolutional}, for example, enhances traditional LSTM by using convolution operations instead of linear ones, achieving notable success in VFP. TrajGRU \cite{shi2017deep} extends GRU by incorporating dynamic, learnable spatial connections, enabling more effective spatiotemporal modeling. Other significant variants include PredRNN \cite{wang2017predrnn}, PredRNN++ \cite{wang2018predrnn++},  Memory in Memory (MIM) \cite{wang2019memory}, and Motion-Aware Unit (MAU) \cite{chang2021mau}. However, sequence models often face challenges such as handling long-term dependencies, vanishing gradient problems, high computational cost, slow training and inference, and susceptibility to error accumulation \cite{ye2022vptr}.

VFP systems have evolved to more advanced techniques leveraging transformers and attention mechanisms, which can capture long-range dependencies and parallelize computation better than previous sequence models. \cite{lin2020self} incorporated the self-attention mechanism into ConvLSTM to capture long-range dependencies in both space and time domains. By integrating the Self-Attention Memory (SAM) unit, the model can better weigh the importance of different spatial and temporal regions, allowing it to focus on more relevant parts of the input data when making predictions. This results in improved performance in tasks such as video prediction, weather forecasting, or any application that requires understanding complex spatiotemporal dynamics. Another significant contribution is the "VPTR: Efficient Transformers for Video Prediction" \cite{ye2022vptr}, which introduces a novel transformer block combining local spatial attention and temporal attention to reduce computational complexity while maintaining high prediction accuracy. The authors propose two models: VPTR-FAR (fully autoregressive) and VPTR-NAR (non-autoregressive), both demonstrating improved performance and reduced error accumulation during inference. Hybrid models combining transformers with other neural network architectures have also been proposed. A notable example is the hybrid Transformer-LSTM model with 3D Separable Convolution for Video Prediction \cite{mathai2024hybrid} which leverages the strengths of both transformers and LSTMs. This model uses the transformer's attention mechanism to capture long-range dependencies and the LSTM's capability to handle temporal sequences, achieving impressive results in video prediction tasks. A key challenge in transformer-based next-frame prediction is the semantic dilution that occurs when the input is split into multiple chunks for MHSA. Furthermore, the commonly used L1 and L2 loss functions, which focus on minimizing the error between the predicted and actual frames, can create a mismatch between the training objective and the model output. This mismatch may introduce unnecessary difficulties in the model's learning process. To the best of our knowledge, these issues have not been explored or addressed.
\section{Method} \label{sec:method}

\subsection{Problem Formulation}
The goal of single-frame VFP is to predict the next frame $I_{t+1}$ using the preceding $M$ frames from a video sequence $\left\{ I_t \right\}_{t=1}^T$, where $I_t \in \mathbb{R}^{H \times W \times C}$ denotes the frame at time $t$, with height $H$, width $W$, and $C$ color channels. The task is modeled as: 
\begin{equation}
    \hat{I}_{T+1} = f_{\theta}(I_{T-M+1}, I_{T-M+2}, \dots, I_T)
\end{equation}
where $f_{\theta}$ maps the sequence of input frames to the predicted next frame $\hat{I}_{T+1}$. $f_{\theta}$ is parameterized by $\theta$.

The prediction error is optimized by minimizing the loss function $\mathcal{L}$, typically defined as the difference between the predicted frame $\hat{I}_{T+1}$ and the ground truth frame $I_{T+1}$. Common choices for $\mathcal{L}$ include the mean squared error (MSE) or mean absolute error (MAE):
\begin{equation}
    % \scriptsize
    \mathcal{L}(\theta) = \frac{1}{HWC} \sum_{i=1}^{H} \sum_{j=1}^{W} \sum_{k=1}^{C} \left( \hat{I}_{T+1}(i, j, k) - I_{T+1}(i, j, k) \right)^2
\end{equation}

The VFP task can then be viewed as solving the following optimization problem:
\begin{equation}
    \theta^* =  \underset{\theta}{\text{argmin}} \, \mathcal{L}(\theta)
\end{equation}
where $\theta^*$ represents the optimal parameters that minimize the prediction error.

\subsection{Transformer-based VFP systems}
Each frame $I_t$ is passed through a CNN-based model $g_\phi$ (e.g., ResNet) to extract a high-dimensional feature representation:
\begin{equation}
    e_t = g_{\phi}(I_t), \quad e_t \in \mathbb{R}^d
\end{equation}
where $e_t$ is the frame embedding, $d$ is dimensionality, and $\phi$ are the CNN parameters.

The embeddings of the last $M$ frames, $E = \{ e_{T-M+1}, e_{T-M+2}, ..., e_T\}$, are input to a Transformer using multi-head self-attention (MHSA) to capture temporal dependencies. For each head $h_i$, the query, key, and value matrices are computed as:
\begin{equation}
    q_t^{h_i} = W_q^{h_i} e_t^{h_i} \quad 
    k_t^{h_i} = W_k^{h_i} e_t^{h_i} \quad 
    v_t^{h_i} = W_v^{h_i} e_t^{h_i}
\end{equation}
where $e_t^{h_i} \in \mathbb{R}^{d_h}$ (with $d_h = \frac{d}{N}$) is the input for each head, and $N$ is the number of heads. This division causes semantic dilution in MHSA.

The attention score between embeddings $e_t$ and $e_{t'}$ for the 
$i$-th head is computed as:
\begin{equation}
    \alpha_{t,t'}^{h_i} = \frac{\exp\left(\frac{q_t^{h_i} \cdot \mathbf{k}_{t'}^{h_i}}{\sqrt{d_h}}\right)} {\sum_{j=1}^{M} \exp\left(\frac{q_t^{h_i} \cdot \mathbf{k}_{j}^{h_i}}{\sqrt{d_h}}\right)}
\end{equation}

The output for the $i$-th head is a weighted sum of the value matrices:
\begin{equation}
    y_t^{h_i} = \sum_{t'=1}^{M} \alpha_{t,t'}^{h_i} v_{t'}^{h_i}
\end{equation}

The outputs from all heads are then concatenated to produce the final output for each embedding $e_t$:
\begin{equation}
    y_t = \left[ y_t^{h_1}; y_t^{h_2}; \dots; y_t^{h_N} \right]
\end{equation}

The output sequence $Y = \{y_1, y_2, ..., y_M\}$ is passed through a decoder network, which generates the predicted frame $\hat{I}_{M+1}$:
\begin{equation}
    \hat{I}_{M+1} = Decoder(Y)
\end{equation}
Typically, the decoder consists of a series of transposed convolutions or upsampling layers that map the final hidden state back to the original frame space.

The model is trained to minimize a loss function, such as the MSE between predicted frame $\hat{I}_{M+1}$ and the ground truth frame $I_{M+1}$:
\begin{equation}
    \mathcal{L}(\theta) = \frac{1}{2}(\hat{I}_{M+1} - I_{M+1})^2
\end{equation}

\subsection{Proposed Method}
\subsubsection{Model Architecture}
We propose a Semantic Concentration VFP (SC-VFP) model (Figure \ref{fig:model_architecture}) to address the limitations of Transformer-based VFP systems. The architecture consists of:

\begin{figure*}[!ht]
    \centering
    \includegraphics[width=1\linewidth]{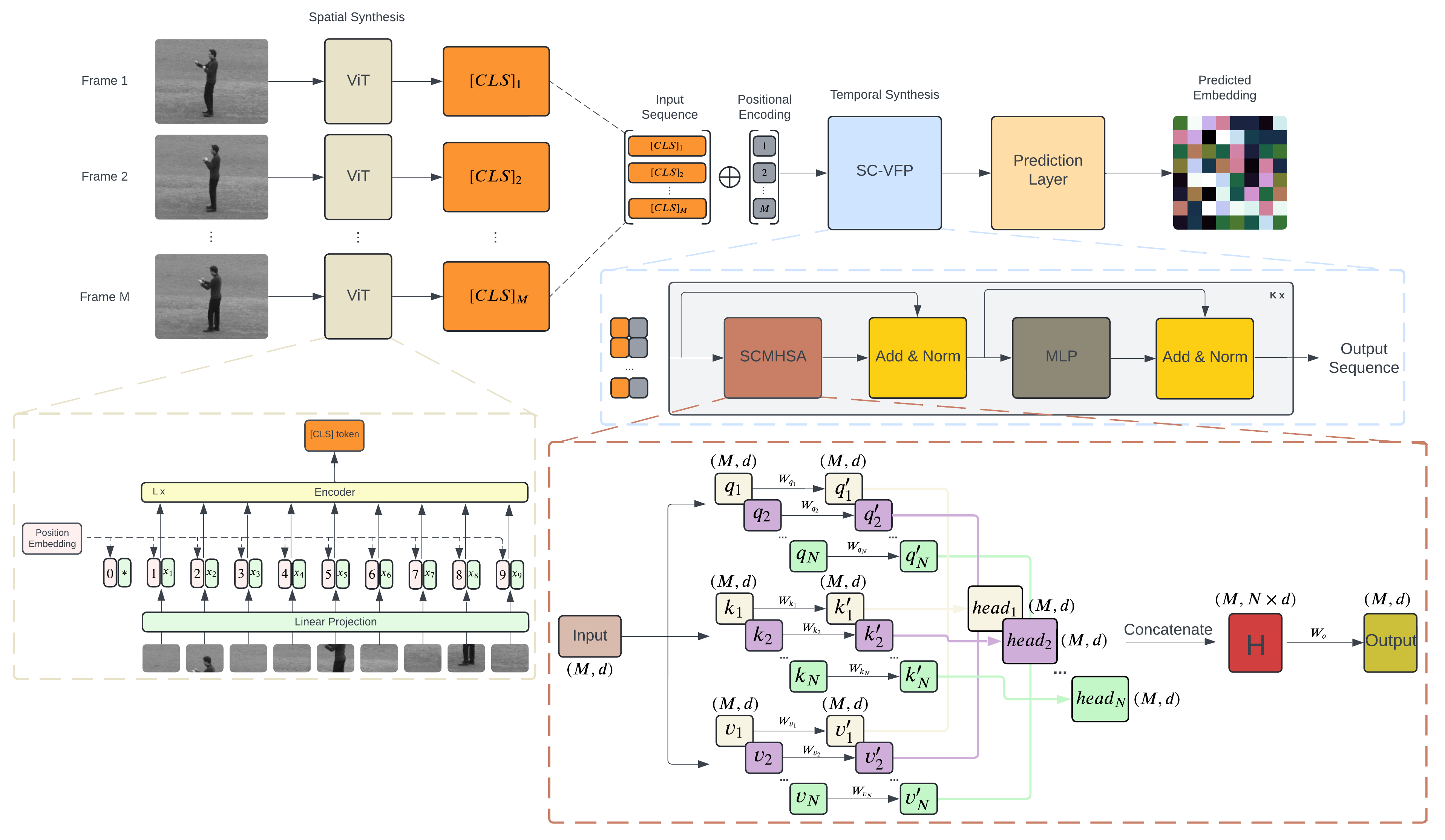}
    \caption{Overview of our proposed network model. We first extract embeddings from $M$ past video frames using the ViT model \cite{dosovitskiy2020image}, utilizing only the classification token [CLS] to represent each frame. Positional encoding is then added, followed by the application of the Semantic Concentration VFP (SC-VFP) module to integrate temporal information. The SC-VFP module includes the SCMHSA block, designed to address the semantic dilution issue in Transformer-based VFP systems. Finally, the processed data is passed through an MLP prediction layer to generate the predicted embedding for the next frame.}
    \label{fig:model_architecture}
\end{figure*}

\begin{enumerate}
    \item \textbf{Embedding Layer:} Maps input frames to a lower-dimensional space via a Vision Transformer (ViT) \cite{dosovitskiy2020image}. A learnable classification token [CLS] aggregates spatial information from the entire input image, representing image frame embeddings.
    \item \textbf{Semantic Concentration VFP (SC-VFP):} Processes embeddings using an encoder-only Transformer, replacing the original MHSA block with our Semantic Concentration MHSA (SCMHSA) block to mitigate semantic dilution. Multiple encoder blocks process temporal information across sequences.
    \item \textbf{Semantic Concentration Multi-head Self-attention (SCMHSA):} UEnhances traditional MHSA by processing the full embedding for each head, yielding richer representations. To manage the increased head dimensions, a learnable projection matrix \( W_o \) retains the most relevant semantic information.
    \item \textbf{Prediction Layer:} Synthesizes temporal and spatial information via a Multi-Layer Perceptron (MLP) to predict the next frame’s embedding.
\end{enumerate}

\subsubsection{Semantic Concentration Multi-Head Self-Attention (SCMHSA)}

To address semantic dilution, SCMHSA enhances standard MHSA by preserving the full semantic content of the input embedding. Unlike traditional MHSA, which splits the embedding $e_t \in \mathbb{R}^d$ into $N$ smaller chunks $e_t^{h_i} \in \mathbb{R}^{d_h = \frac{d}{N}}$ (for $N$ heads), SCMHSA feeds the complete embedding to each attention head, enabling holistic processing and mitigating semantic loss.

Building on the standard Transformer architecture described earlier, SCMHSA introduces the following key modifications:
\begin{enumerate}
    \item \textbf{Full Embedding for Each Head}: Each attention head $h_i$ processes the entire embedding $e_t \in \mathbb{R}^d$, avoiding division:
    \begin{equation}
        q_t^{h_i} = W_q^{h_i} e_t \quad
        k_t^{h_i} = W_k^{h_i} e_t \quad
        v_t^{h_i} = W_v^{h_i} e_t 
    \end{equation}
    $W_q^{h_i}, W_k^{h_i}, W_v^{h_i} \in \mathbb{R}^{d \times d'_h}$ are learned projection matrices, with $d'_h$ representing the head-specific projection dimensionality.
    \item \textbf{Learnable Projection:} The outputs of all heads are concatenated:
    \begin{equation}
        y_t = \left[ y_t^{h_1}; y_t^{h_2}; \dots; y_t^{h_N} \right] \in \mathbb{R}^{h \times d'_h}
    \end{equation}
    To handle the increased dimensionality and retain key semantics, a learnable projection matrix, $W_o \in \mathbb{R}^{(h \times d'_h) \times d}$, reduces the dimensionality back to $d$ while preserving the most relevant semantic information:
    \begin{equation}
        y_t^{final} = W_o y_t
    \end{equation}

    % This final output $y_t^{final}$ retains the full semantic meaning of the input embedding, enriched by the MHSA mechanism, and is ready to be used for subsequent layers or the final prediction task.
\end{enumerate}

SCMHSA provides:
\begin{itemize}
    \item Semantic Integrity: Each head retains the full semantic context, avoiding dilution.
    \item Holistic Representation: All heads process complete embeddings for richer understanding.
    \item Dimensionality Management:  The learnable matrix $W_o$ filters irrelevant features, preserving key information.
\end{itemize}

\subsubsection{Prediction Layer}
The SC-VFP module outputs a sequence of vectors $Y = \{y^{final}_1, y^{final}_2, ..., y^{final}_M \}$, where $M$ is the number of previous frames, and $y^{final}_i \in \mathbb{R}^d$ represents the processed embedding of the $t$-th frame, enriched with the full semantic information. The Prediction Layer (implemented as an MLP), takes the output from SC-VFP module and generates a vector $\hat{e}_{M+1} \in \mathbb{R}^d$ that predicts the embedding of the next frame in the sequence.

The prediction layer is designed to translate the temporal and spatial information captured by SC-VFP into an accurate prediction of the next video frame's embedding.

\subsubsection{Our Loss Function}
In existing Transformer-based VFP systems, the model predicts the next frame's embedding, but the loss is computed on the reconstructed frame \cite{lin2020self}, \cite{ning2023mimo}, \cite{girdhar2021anticipative}, \cite{ye2022vptr}, which can introduce learning discrepancies. To address this, we propose a new loss function that directly optimizes the predicted embedding and enhances the effectiveness of the SCMHSA module.

The proposed loss function $\mathcal{L}$ consists of two main components:
\begin{enumerate}
    \item \textbf{Embedding MSE loss:} This component directly measures the error between the true embedding $e_t$ and the predicted embedding $\hat{e_t}$. This helps to ensure the model is trained to produce accurate embedding of the next frame.
    \begin{equation}
        \mathcal{L}_{\text{MSE}} = \text{MSE}(e_t, \hat{e}_t) = \left\lVert e_t - \hat{e}_t \right\rVert^2
    \end{equation}
    \item \textbf{Semantic Similarity loss:} Given that SCMHSA allows each head to receive the full input embedding, it is crucial to ensure that the heads capture distinct, non-overlapping semantic information. The semantic similarity loss achieves this by penalizing heads that produce similar outputs.

    For each pair of heads $h_i$ and $h_j$ (where $i \neq j$), this component computes the row-wise cosine similarity as the average angle distance between $h_i$ and $h_j$. The sum of these row-wise cosine similarities is averaged across all pairs of heads and all rows in the head vector. The Semantic Similarity loss can be expressed as:
    \begin{equation}
        \mathcal{L}_{\text{SS}} = \frac{1}{N(N-1)} \sum_{i=1}^{N-1} \sum_{j=i+1}^{N} \frac{\sum_{k=1}^{M} \left|\cos(h_{i,k}, h_{j,k})\right|}{M}
    \end{equation}
    where $N$ is the number of heads, $M$ is the number of rows in each head vector (which is also the length of the input sequence), and $h_{i,k}$ is the $k$-th row of head $h_i$.
\end{enumerate}

The total loss function $\mathcal{L}$ is a weighted sum of the two components:
\begin{equation}
    \mathcal{L} = \mathcal{L}_{MSE} + \lambda \mathcal{L}_{SS}
\end{equation}
where $\lambda$ is a hyperparameter that controls the trade-off between the MSE loss and the Semantic Similarity loss.
\section{Experiments} \label{sec:experiments}

\subsection{Datasets}
We conducted our experiments using four datasets: KTH, UCSD Pedestrian, UCF Sports, and Penn Action. Each training instance comprised six frames: five input frames and a sixth as the label. Instead of selecting five consecutive frames as input, we chose one frame for every five to avoid similarity issues with consecutive frames. This approach also reduces the amount of training data that needs to be processed. All input frames were resized to $224 \times 224$ for compatibility with the ViT model \cite{dosovitskiy2020image}.

\subsubsection{KTH \cite{schuldt2004recognizing}}
The KTH dataset is widely used for human action recognition, containing 600 video sequences with a resolution of $160 \times 120$, and recorded at 25 fps. For our experiments, we use the walking and running classes. 

\subsubsection{UCSD Pedestrian \cite{5539872}} The UCSD Pedestrian dataset features video footage from a stationary, elevated camera capturing pedestrian walkways to identify abnormal events. The dataset has two subsets: Peds1, with 34 training and 36 testing videos of people walking towards or away from the camera, and Peds2, with 16 training and 12 testing videos of pedestrians moving parallel to the camera. Each clip has $200$ black-and-white frames at a $238 \times 158$ resolution.

\subsubsection{UCF Sports \cite{soomro2015action}}
The UCF Sports dataset contains 150 video sequences, each with a resolution of $720 \times 480$, and encompasses 10 action classes. This dataset is useful for studying human actions in sports contexts, offering diverse scenarios and movements.

\subsubsection{Penn Action \cite{zhang2013actemes}}
The Penn Action dataset contains 2,326 video sequences with a resolution of $640 \times 480$, covering 15 action classes, such as baseball pitching, pull-ups, and guitar strumming. It is valuable for action recognition tasks due to its diverse range of physical activities and detailed labeling of dynamic motions.

\subsection{Metrics}
Since our approach operates within the embedding space rather than the pixel space, common VFP metrics such as LPIPS \cite{zhang2018unreasonable} and SSIM are not applicable. Instead, we employ PSNR and MSE as our evaluation metrics. While PSNR is conventionally calculated using the pixel values, we modify it to utilize the embedding values.

\subsection{Implementation Details}
The proposed model was implemented using PyTorch\footnote{Torch version 2.3.0, Torchvision version 0.18.0, Torchaudio version 2.3.0 }. Training and evaluation were conducted on an NVIDIA A100 40GB GPU. The model's architecture consisted of 6 encoder blocks, each with 6 attention heads and a frame embedding dimension of 768. The sequence length was set to 5. For optimization, we used the AdamW optimizer \cite{loshchilov2017decoupled} with a learning rate of 1e-4. The batch size was set to 32, and the model was trained for 25 epochs. The dataset was split into training, validation, and test sets with a ratio of 0.7, 0.15, and 0.15, respectively. To ensure reproducibility, we fixed the random seed at 2023.

\subsection{Results}
\begin{figure*}[!ht]
    \centering
    \includegraphics[width=\textwidth]{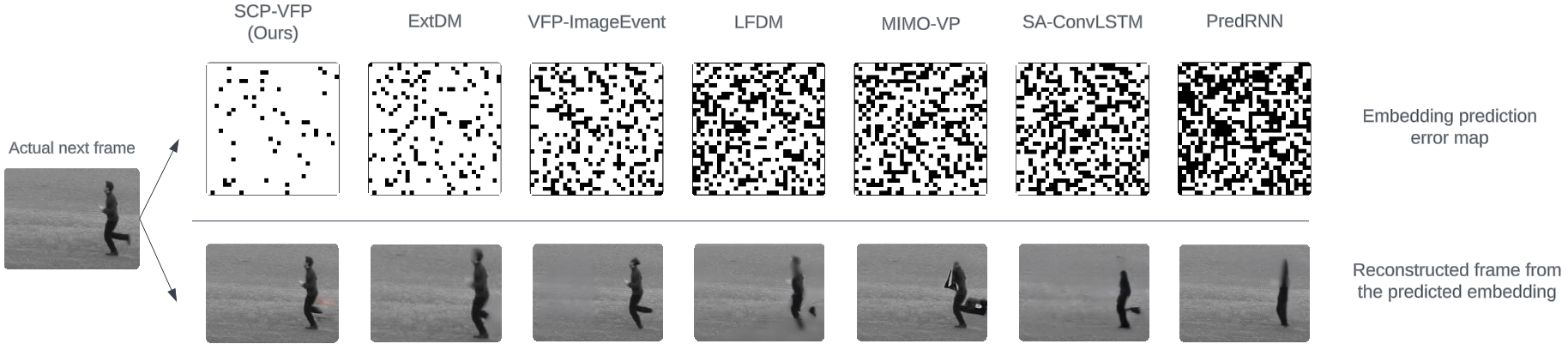}
    \caption{Qualitative comparison of methods on an example frame from KTH dataset. The top shows the error map of the predicted embedding and the actual embedding. The darker the error map appears, the worse the prediction is to the ground truth. The bottom shows the reconstructed frame using the predicted embedding.}
    \label{fig:difference_maps}
\end{figure*}

\begin{figure}[!ht]
    \centering
    \includegraphics[width=\columnwidth]{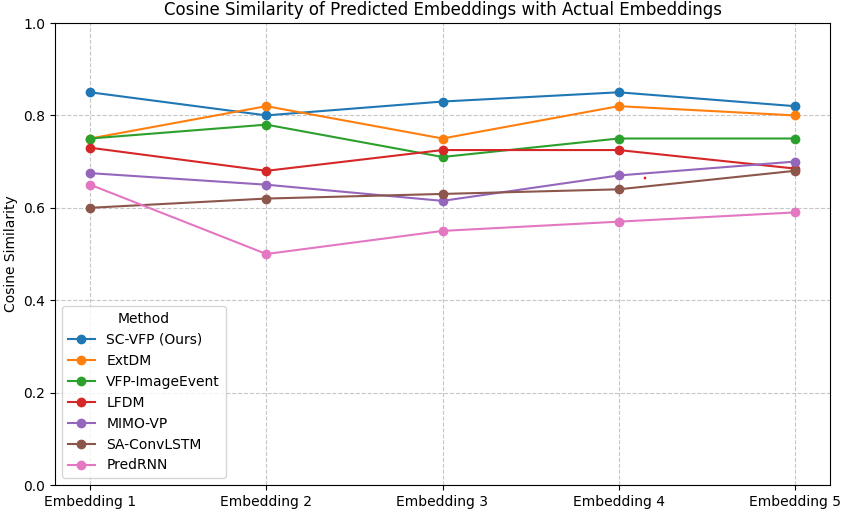}
    \caption{Consine similarity between the first five predicted embeddings and ground truth embeddings of all methods on KTH dataset. A higher cosine similarity indicates that the prediction is closer to the ground truth.}
    \label{fig:cosine}
\end{figure}

We present the results of our proposed method, comparing it with the relevant and latest next-frame predictors that we know: PredRNN \cite{wang2017predrnn}, SA-ConvLSTM \cite{lin2020self}, MIMO-VP \cite{ning2023mimo}, LFDM \cite{ni2023conditional}, VFP-ImageEvent \cite{zhu2024video}, and ExtDM \cite{zhang2024extdm} across the four datasets previously mentioned. All methods were evaluated on the test set of these four datasets. The quantitative results and comparisons are summarized in Table \ref{tab:quantitative}. In addition to the quantitative comparison, we present a qualitative comparison in Figure \ref{fig:difference_maps}. Unlike conventional next-frame prediction models, which directly predict the next frame, our proposed method predicts the embedding of the next frame instead. The ground truth embedding of the next frame is generated using ViT \cite{dosovitskiy2020image}. In Figure \ref{fig:difference_maps} we visualize the error maps of the predicted embedding and the ground truth embedding for each method. The qualitative results demonstrate how closely the predicted embeddings from our method align with the ground truth. The darker the error map appears, the worse the prediction is to the ground truth. We also compare the cosine similarity between the embedding predicted by our method (SC-VFP) and the ground truth embedding, together with other methods in Figure \ref{fig:cosine}.

\begin{table*}[!ht]
\centering
\begin{tabular}{|c|c|cc|cc|cc|cc|}
\hline
\multirow{2}{*}{\textbf{Method}} & \multirow{2}{*}{\textbf{\#Params}} & \multicolumn{2}{c|}{\textbf{KTH}} & \multicolumn{2}{c|}{\textbf{UCSD}} & \multicolumn{2}{c|}{\textbf{UCF Sports}} & \multicolumn{2}{c|}{\textbf{Penn Action}} \\
\cline{3-10}
 & & MSE $\downarrow$ & PSNR $\uparrow$ & MSE $\downarrow$ & PSNR $\uparrow$ & MSE $\downarrow$ & PSNR $\uparrow$ & MSE $\downarrow$ & PSNR $\uparrow$ \\
\hline
PredRNN & 13.8M & 359.82 & 22.57 & 350.82 & 22.68 & 507.08 & 21.08 & 633.99 & 20.11 \\
SA-ConvLSTM & 10.5M & 248.36 & 24.18 & 255.91 & 24.05 & 380.26 & 22.33 & 425.68 & 21.84 \\
MIMO-VP & 30.2M & 154.91 & 26.23 & 154.91 & 26.23 & 217.81 & 24.75 & 247.79 & 24.19 \\
LFDM & 54.2M & 133.07 & 26.89 & 168.3 & 25.87 & 197.73 & 25.17 & 278.66 & 23.68 \\
VFP-ImageEvent & 83.6M & \textbf{49.9} & \textbf{31.15} & 100.71 & 28.1 & 134.3 & 26.84 & 185.41 & 25.88 \\
ExtDM & 80.5M & 95.3 & 28.34 & 125.92 & 27.13 & 110.17 & 27.71 & 199.56 & 25.13 \\
SC-VFP (Ours) & 42.7M & 79.81 & 29.11 & \textbf{86.71} & \textbf{28.75} & \textbf{79.63} & \textbf{29.12} & \textbf{109.92} & \textbf{27.72} \\
\hline
\end{tabular}
\caption{Quantitative comparison of methods across different datasets. Bold indicates the highest performance.}
\label{tab:quantitative}
\end{table*}

\subsubsection{Results on KTH}
KTH is the only dataset where our method exhibited inferior performance compared to others. Specifically, SC-VFP achieved a PSNR score that was $7.01\%$ lower than the best-performing method and an MSE score that was $59.94\%$ lower (Table \ref{tab:quantitative}). This can be attributed to the relatively small size of the KTH dataset compared to the other three datasets used in our experiments. As a result, the issue of semantic dilution is less pronounced during training and testing on this dataset. In contrast, larger datasets contain more diverse semantics, increasing the potential impact of semantic dilution, which may result in more noticeable distinctions in performance. 
% Figure \ref{fig:KTH} presents a qualitative comparison between our SC-VFP model and other models. A certain degree of blurriness can be observed in the embedding predictions generated by our model compared to those produced by the best-performing method (MIMO-VP).

\subsubsection{Result on UCSD Pedestrian}
On the UCSD dataset, our SC-VFP method outperformed all others, achieving the lowest MSE of 86.71 and the highest PSNR of 28.75. These results represent an $16.14\%$ improvement in MSE and a $2.59\%$ improvement in PSNR over the next best method (VFP-ImageEvent) (Table \ref{tab:quantitative}). 
% Figure \ref{fig:UCSD} provides the qualitative comparison, reflecting the effectiveness of our method in predicting the embedding of the next frame.
% \begin{figure}[!ht]
%     \centering
%     \includegraphics[width=0.85\linewidth]{images/UCSD.png}
%     \caption{Embedding prediction examples on UCSD Pedestrian dataset.}
%     \label{fig:UCSD}
% \end{figure}

\subsubsection{Result on UCF Sports}
On the UCF Sports dataset, which is larger and more complex than KTH and UCSD, we observe a more significant performance gap between SC-VFP and others. Specifically, the MSE was reduced by 38.3\%, and the PSNR increased by 4.84\% over the nearest competitor (ExtDM). These results further validate our hypothesis, suggesting that addressing semantic dilution is crucial for handling the more complex motion patterns found in larger datasets. 
% Figure \ref{fig:UCF} provides the qualitative comparison on UCF Sports.
% \begin{figure}[!ht]
%     \centering
%     \includegraphics[width=0.9\linewidth]{images/UCF Sports.png}
%     \caption{Embedding prediction examples on UCF Sports dataset.}
%     \label{fig:UCF}
% \end{figure}

\subsubsection{Result on Penn Action}
Evaluating on the largest dataset (Penn Action), we observe a continued trend of improvement on larger and more complex datasets. SC-VFP once again establishes a substantial performance gap, outperforming the second-best method by $68.71\%$ in MSE and $6.63\%$ in PSNR. This demonstrates the scalability and robustness of our approach, confirming the suitability of SC-VFP for advanced video prediction tasks where fine-grained semantic information is crucial.
% Figure \ref{fig:Penn} provides the qualitative results.
% \begin{figure}[!ht]
%     \centering
%     \includegraphics[width=0.85\linewidth]{images/Penn Action.png}
%     \caption{Embedding prediction examples on Penn Action dataset.}
%     \label{fig:Penn}
% \end{figure}

\subsection{Ablation Study}
\subsubsection{Parameter Analysis}
Our proposed SCMHSA model (42.7M parameters) introduces an increase in parameter count compared to the original Transformer-based model (31.4M parameters) due to its complete embedding processing in each attention head, which mitigates semantic dilution. This design results in approximately 1.35x more parameters than the baseline, enabling richer semantic representations and improving predictive accuracy, particularly on complex datasets. While our model has a higher parameter count relative to the original Transformer, experiments show that the additional parameters in SCMHSA significantly enhance the model’s ability to capture complex spatiotemporal dynamics, leading to notable improvements in MSE and PSNR metrics across diverse datasets, as demonstrated in the next section. The parameter counts of our method and the original Transformer-based method are shown in Table \ref{tab:params_compare}.

\begin{table}[!ht]
    \centering
    \begin{tabular}{|c|c|}
    \hline
    \textbf{Method} & \textbf{\#Params} \\
    \hline
    With SCMHSA & 42.7M \\
    Without SCMHSA & 31.4M \\
    \hline
    \end{tabular}
    \caption{Comparison of parameter count between SC-VFP with and without SCMHSA.}
    \label{tab:params_compare}
\end{table}

\subsubsection{Performance Analysis}
\begin{figure}[!ht]
    \centering
    \includegraphics[width=1\columnwidth]{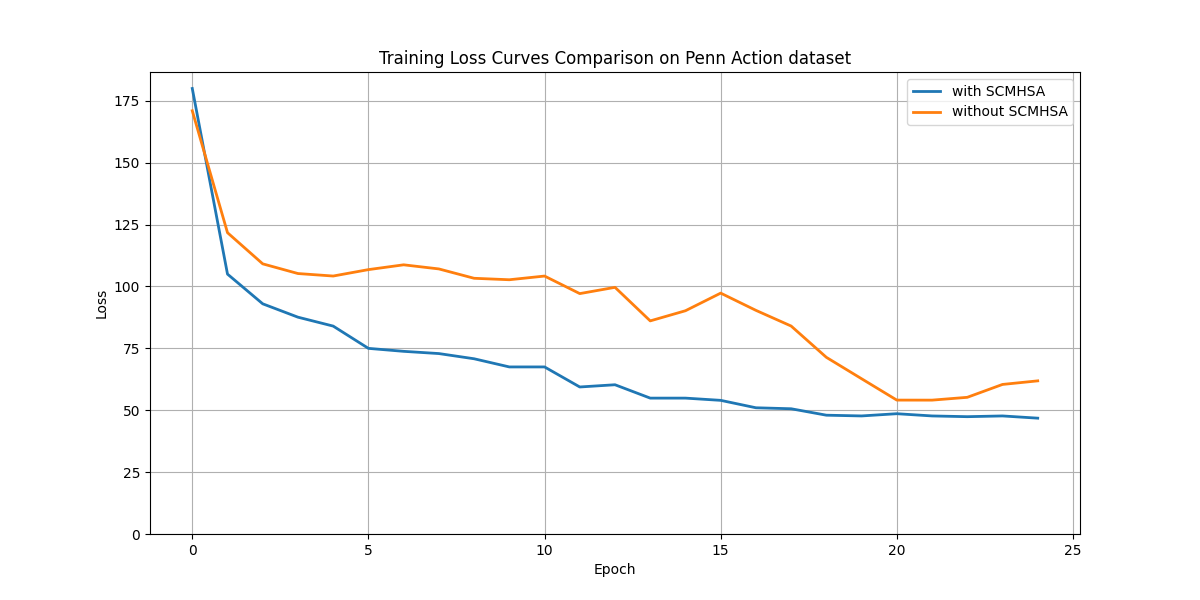}
    \caption{Convergence speed of SC-VFP with and without SCMHSA on the training set of Penn Action dataset. SC-VFP with SCMHSA not only achieves a lower loss but also converges more quickly}
    \label{fig:ablation}
\end{figure}

To verify the contributions of SCMHSA and the Semantic Similarity loss modules, we performed an ablation study on the test set of four datasets (KTH, UCSD, UCF Sports, Penn Action). Table \ref{tab:ablation_SCMHSA_combined} demonstrates that excluding SCMHSA leads to varied impacts across different datasets. On the KTH dataset, SC-VFP without SCMHSA observes a better performance ($0.45\%$ lower on MSE and $0.07\%$ higher on PSNR). In contrast, on the UCSD dataset, incorporating SCMHSA gains better performance, which decreases the MSE by $28.87\%$ and enhances the PSNR by $3.97\%$. The effect is even more pronounced on the UCF Sports dataset, where the exclusion of SCMHSA significantly worsens the MSE by $45.29\%$ and PSNR by $9.89\%$. Finally, on the Penn Action dataset, the inclusion of SCMHSA improves the MSE and PSNR values by $35.71\%$ and $6.92\%$, respectively. These results underscore the critical role of the SCMHSA mechanism in mitigating the semantic dilution problem, thereby contributing significantly to accurate embedding predictions, particularly in larger and more complex datasets. Figure \ref{fig:ablation} illustrates the convergence trend of SC-VFP with and without SCMHSA on the training set of the Penn Action dataset. The results indicate that SC-VFP with SCMHSA not only achieves a lower loss but also converges more quickly compared to the variant without SCMHSA.

\setlength{\tabcolsep}{1mm} % Adjust the space between columns

\begin{table}[!ht]
\centering
\begin{tabular}{|c|cc|cc|}
\hline
\multirow{2}{*}{\textbf{Method}} & \multicolumn{2}{c|}{\textbf{KTH}} & \multicolumn{2}{c|}{\textbf{UCSD}} \\
\cline{2-5}
 & MSE $\downarrow$ & PSNR $\uparrow$ & MSE $\downarrow$ & PSNR $\uparrow$ \\
\hline
With SCMHSA & 79.81 & 29.11 & 86.71  & 28.75 \\
Without SCMHSA & 79.45 & 29.13 & 111.71 & 27.65 \\
\hline
\multicolumn{5}{|c|}{} \\[-0.8em]
\hline
\multirow{2}{*}{\textbf{Method}} & \multicolumn{2}{c|}{\textbf{UCF Sports}} & \multicolumn{2}{c|}{\textbf{Penn Action}} \\
\cline{2-5}
 & MSE $\downarrow$ & PSNR $\uparrow$ & MSE $\downarrow$ & PSNR $\uparrow$ \\
\hline
With SCMHSA & 79.63 & 29.12 & 109.92 & 27.72 \\
Without SCMHSA & 145.57 & 26.5 & 171.03 & 25.8 \\
\hline
\end{tabular}
\caption{Ablation results on SCMHSA module for all datasets. The first section represents results on KTH and UCSD datasets, while the second section represents results on UCF Sports and Penn Action datasets.}
\label{tab:ablation_SCMHSA_combined}
\end{table}

The ablation results for the Semantic Similarity Loss (SSL) (Table \ref{tab:ablation_SSL_combined} reveals that SSL plays an even more crucial role in enhancing performance. Even with the inclusion of SCMHSA, the omission of SSL leads to significantly worse performance in MSE across four datasets, with the amount of $26.4\%$, $48.2\%$, $128.5\%$, and $96.8\%$, respectively. Similarly, the PSNR performance decreases by $3.27\%$, $6.32\%$, $14.06\%$, and $11.86\%$ across these four  datasets. These findings underscore the substantial impact of SSL on improving the model’s accuracy, even in the presence of other enhancements like SCMHSA.

\setlength{\tabcolsep}{1mm} % Reduce the space between columns

\begin{table}[!ht]
\centering
\begin{tabular}{|c|cc|cc|}
\hline
\multirow{2}{*}{\textbf{Method}} & \multicolumn{2}{c|}{\textbf{KTH}} & \multicolumn{2}{c|}{\textbf{UCSD}} \\
\cline{2-5}
 & MSE $\downarrow$ & PSNR $\uparrow$ & MSE $\downarrow$ & PSNR $\uparrow$ \\
\hline
With SSL & 79.81 & 29.11 & 86.71  & 28.75 \\
Without SSL & 101.88 & 28.05 & 130.34 & 26.98 \\
\hline
\multicolumn{5}{|c|}{} \\[-0.8em]
\hline
\multirow{2}{*}{\textbf{Method}} & \multicolumn{2}{c|}{\textbf{UCF Sports}} & \multicolumn{2}{c|}{\textbf{Penn Action}} \\
\cline{2-5}
 & MSE $\downarrow$ & PSNR $\uparrow$ & MSE $\downarrow$ & PSNR $\uparrow$ \\
\hline
With SSL & 79.63 & 29.12 & 109.92 & 27.72 \\
Without SSL & 183.27 & 25.5 & 215.32 & 24.8 \\
\hline
\end{tabular}
\caption{Ablation results on Semantic Similarity Loss (SSL) for all datasets. The first section presents results on KTH and UCSD datasets, while the second section presents results on UCF Sports and Penn Action datasets.}
\label{tab:ablation_SSL_combined}
\end{table}

\section{Conclusion} \label{sec:conclusion}
In this paper, we have addressed the issue of semantic dilution in Transformer-based video frame prediction (VFP) systems by introducing a novel solution called Semantic Concentration VFP (SC-VFP). Our approach leverages the Semantic Concentration Multi-head Self-attention (SCMHSA) mechanism and a new loss function designed to align more effectively with VFP outputs. This loss function enhances the ability of high-dimensional attention heads to capture unique, non-overlapping semantics. While SC-VFP demonstrated inferior performance on the smaller dataset (KTH), it achieved state-of-the-art results on three larger datasets (UCSD, UCF Sports, and Penn Action). These results underscore the suitability of SC-VFP for larger datasets that contain more diverse semantics, where the impact of semantic dilution is more pronounced and the need for fine-grained semantic information is critical.
{
    \small
    \bibliographystyle{ieeenat_fullname}
    \bibliography{main}
}

% WARNING: do not forget to delete the supplementary pages from your submission 
% \input{sec/X_suppl}

\end{document}